\documentclass{article}
\usepackage{ijcai19}

\usepackage{times}
\usepackage{soul}
\usepackage{url}
\usepackage[hidelinks]{hyperref}
\usepackage[utf8]{inputenc}
\usepackage[small]{caption}
\usepackage{amsmath}
\usepackage{booktabs}
\usepackage{epsfig}
\usepackage{graphicx}
\usepackage{amsmath}
\usepackage{amssymb}
\usepackage{subcaption}
\usepackage{lipsum}

\usepackage{manyfoot}
\SetFootnoteHook{\hspace*{-1.8em}}
\DeclareNewFootnote{bl}[gobble]
\title{MSR: Multi-Scale Shape Regression for Scene Text Detection}

\author{
Chuhui Xue$^1$ \and
Shijian Lu$^1$ \And
Wei Zhang$^2$
\affiliations
$^1$School of Computer Science and Engineering, Nanyang Technological University\\
$^2$School of Control Science and Engineering, Shandong University\\
\emails
xuec0003@e.ntu.edu.sg,
shijian.lu@ntu.edu.sg,
davidzhang@sdu.edu.cn
}

\begin{document}

\maketitle

\begin{abstract}
State-of-the-art scene text detection techniques predict quadrilateral boxes that are prone to localization errors while dealing with straight or curved text lines of different orientations and lengths in scenes. This paper presents a novel multi-scale shape regression network (MSR) that is capable of locating text lines of different lengths, shapes and curvatures in scenes. The proposed MSR detects scene texts by predicting dense text boundary points that inherently capture the location and shape of text lines accurately and are also more tolerant to the variation of text line length as compared with the state of the arts using proposals or segmentation. Additionally, the multi-scale network extracts and fuses features at different scales which demonstrates superb tolerance to the text scale variation. Extensive experiments over several public datasets show that the proposed MSR obtains superior detection performance for both curved and straight text lines of different lengths and orientations.
\end{abstract}

\section{Introduction}
Automated detection of various texts in scenes has attracted increasing interests in recent years due to its growing demands in many real-world applications such as image search, autonomous driving, etc. With the advance of deep neural networks (DNNs), a number of DNN based scene text detection systems have been reported and many have achieved very promising detection performance. Different scene text detection approaches have been explored which treat scene texts as one specific type of object and adapt various generic object techniques for the scene text detection task.

State-of-the-art scene text detection techniques still suffer from two typical constraints. The first is inaccurate localization due to the specific text line shapes, typically thin, having different lengths and sometimes being curved in different ways. Specifically, proposal based techniques are often at a loss in selecting anchors of appropriate aspect ratios for dealing with text lines of different lengths. Segmentation based techniques often introduce large regression errors as pixels lying around the centre of long text lines are very far from the vertices of quadrilateral regression boxes. In addition, most existing techniques generate rectangular or quadrilateral localization boxes which often include undesired image background while dealing with various curved text lines in scenes. The second is unreliable detection while dealing with texts of abnormal sizes in images. Scene texts usually have much larger scale variations as compared with generic objects \cite{singh2018analysis}, e.g. the scale ratio between the largest and the smallest texts is up to 230 times for images in the MSRA-TD500 \cite{yao2012detecting}. The large scale variation often leads to miss detection for ultra-small text instances or broken/partial detection for ultra-large ones. 

\begin{figure*}[t!]
  \centering
  \includegraphics[width=.95\linewidth]{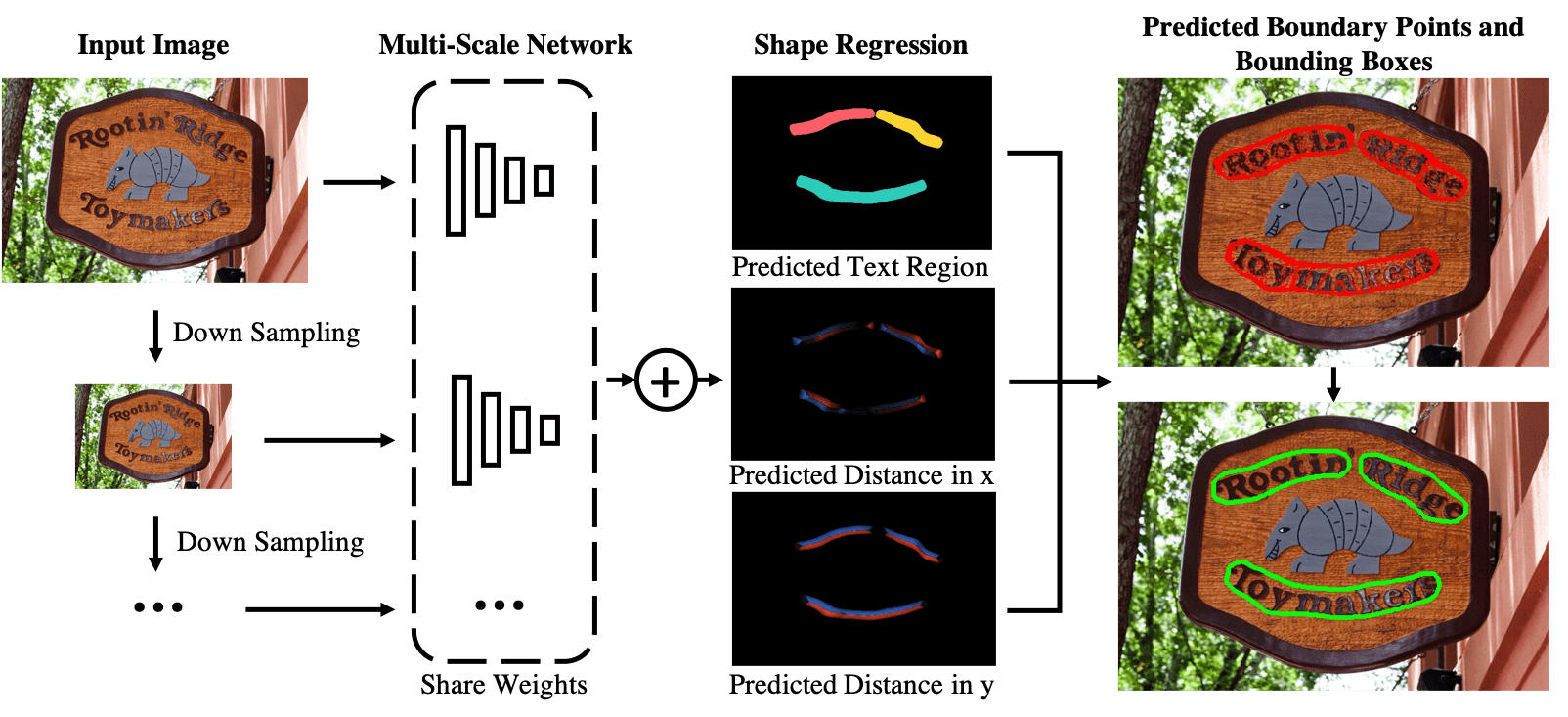}
  \caption{The framework of the proposed technique: An image and its down-scaled are fed to the multi-scale shape regression network (MSR) as input. The MSR employs multiple network channels to extract and fuse features at different scales to predict the central text regions, the distances from the central text regions to the text boundaries and dense text boundary points. Scene texts of different orientations, shapes and lengths are located by a concave polygon that encloses all boundary points of each text instance.
  }
  \label{fig2}
\end{figure*}
We design an innovative multi-scale shape regression network (MSR) that addresses both constraints at one go as illustrated in Fig. \ref{fig2}. Different from existing techniques, the proposed MSR regresses text pixels to the nearest text boundary points and locates scene texts by linking up the regressed text boundary points. It can thus detect scene texts of different orientations and shapes more accurately as compared with most existing techniques which produce quadrilateral vertices and often include undesired image background. In addition, MSR can locate text lines of different lengths more accurately because regressing to the nearest boundary points introduces much less regression errors as compared with regressing to the quadrilateral vertices especially for long text lines. Further, a multi-scale network is designed which empowers MSR with better tolerance to the large scale variation of texts in scenes. Specifically, it employs multiple network channels to extract and fuse features at different scales which leads to more robust detection in the presence of large text size variations. Experiments over several public datasets show that the proposed MSR is broadly applicable and achieves superior detection performance for scene texts with different orientations, shapes and lengths.

The contributions of this work are threefold. First, it proposes a novel shape regression technique to predict dense text boundary points with which scene texts of different orientations, shapes and lengths can be located accurately. Second, it proposes a multi-scale network that employs multiple network channels to extract and fuse features at different scales and demonstrates great tolerance to the large text scale variation. Third, it develops an end-to-end trainable system that achieves superior scene text detection performance over a number of public datasets with scene texts of different orientations, lengths and shapes.

\section{Related Work}

\subsubsection{Scene Text Detection} 
The earlier works \cite{tian2015text,yao2012detecting} detect characters/words using various hand-crafted features. With the fast development of deep neural networks, a lot of CNN-based scene detection approaches have been proposed. One typical approach adapts generic object detection techniques which either leverage text-specific proposals or default boxes \cite{tian2016detecting} or follow the DenseBox \cite{huang2015densebox} idea by first extracting text regions and then regressing each text pixel to the vertices of localization boxes \cite{zhou2017east,he2017deep,xue2018accurate}. In addition, some work treats scene text detection as a text segmentation problem \cite{deng2018pixellink}, which predicts pixel-level text feature map and localizes text instances by segmenting text regions directly from the text feature map. 

The recent scene text detection research predicts quadrilateral boxes of different orientations for multi-oriented text lines which still faces various problems while dealing with text lines of different shapes and lengths. Our proposed shape regression network instead predicts distances from text pixels to the nearest text boundary. It generates dense text boundary points that can be linked up to locate scene texts of very different shapes and lengths accurately.

\subsubsection{Large Scale Variation}
Two typical approaches have been explored in the era of deep learning. The first approach exploits features that are extracted at multiple network layers of different depths, instead of just using features from the last network layer which often carries large-scale global information only. For example, FPN \cite{lin2017feature} adopts a top-down network structure and treats it as a feature pyramid to make predictions at different network layers. U-Net \cite{ronneberger2015u} extracts features from different stages of the backbone network and fuses them by up-sampling features to the same scale. The other approach employs images of different scales in network training. One typical way is to make predictions at images of different scales and then combines them as the final predictions \cite{hao2017scale}. In addition, \cite{singh2018analysis} attempt to predict objects at specific scales where the input image is resized to the corresponding resolutions, e.g. predicting smaller objects at higher resolutions of the input image. 

Our proposed multi-scale network deals with large scale variations by marrying the merits of the two state-of-the-art approaches. Specifically, it designs a new network structure that employs multiple network channels to extract and fuse features at different network stages from images of different scales simultaneously. Experiments show its superior tolerance to large object scale variations.

\begin{figure*}[t!]
  \centering
  \includegraphics[width=\linewidth]{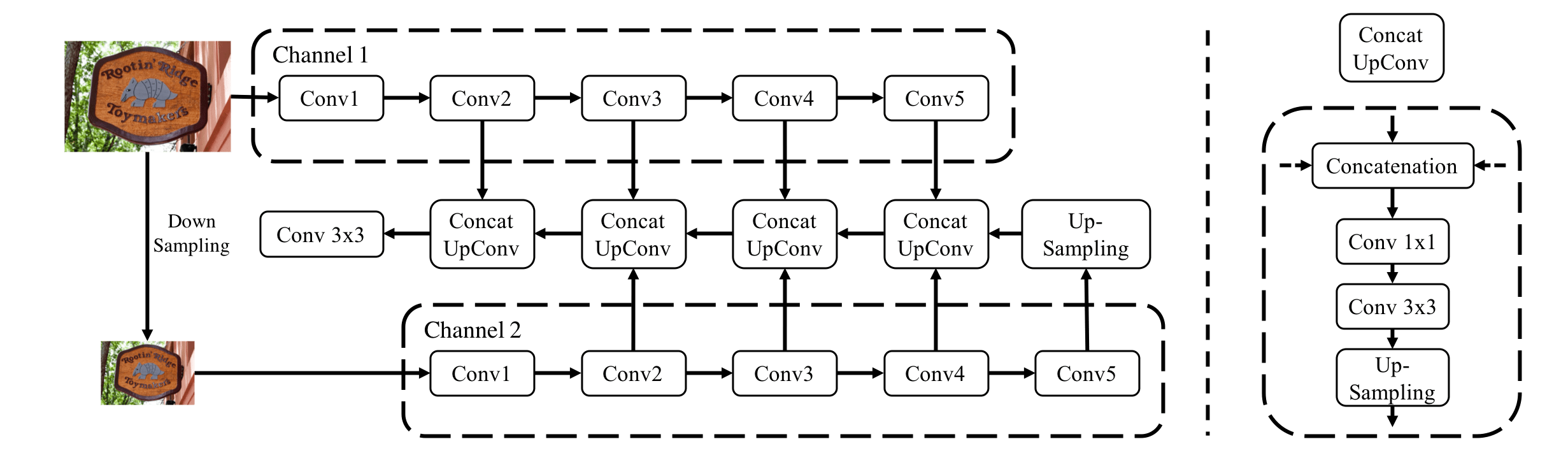}
  \caption{Structure of proposed multi-scale network (for two-scale case): Features extracted from layers \textit{Conv2} - \textit{Conv5} of two network channels are fused, where features of the same scale are fused by a \textit{Concat UpConv} as illustrated and features from the deepest layer of the lower-scale channel are up-sampled to the scale of the previous layer for fusion.
  }
  \label{fig3}
\end{figure*}

\section{Methodology}
We propose a novel multi-scale shape regression network for accurate detection of scene texts of different orientations, shapes and lengths as illustrated in Fig. \ref{fig2}. A \textbf{Multi-Scale Network} is designed to extract and fuse features from images of different scales. The fused features are then fed to a \textbf{Shape Regression} module to detect central text regions and predict distances from each central text region pixel to its nearest text boundary. This produces a set of dense text boundary points (the red points in the image in the top right corner of Fig. \ref{fig2}) that can be linked up to produce polygon localization boxes (the green polygon in the image in the bottom right corner).

\subsection{Shape Regression for Scene Text Localization}
\begin{figure}[t]
\begin{subfigure}{.16\textwidth}
  \centering
  \includegraphics[width=.95\linewidth]{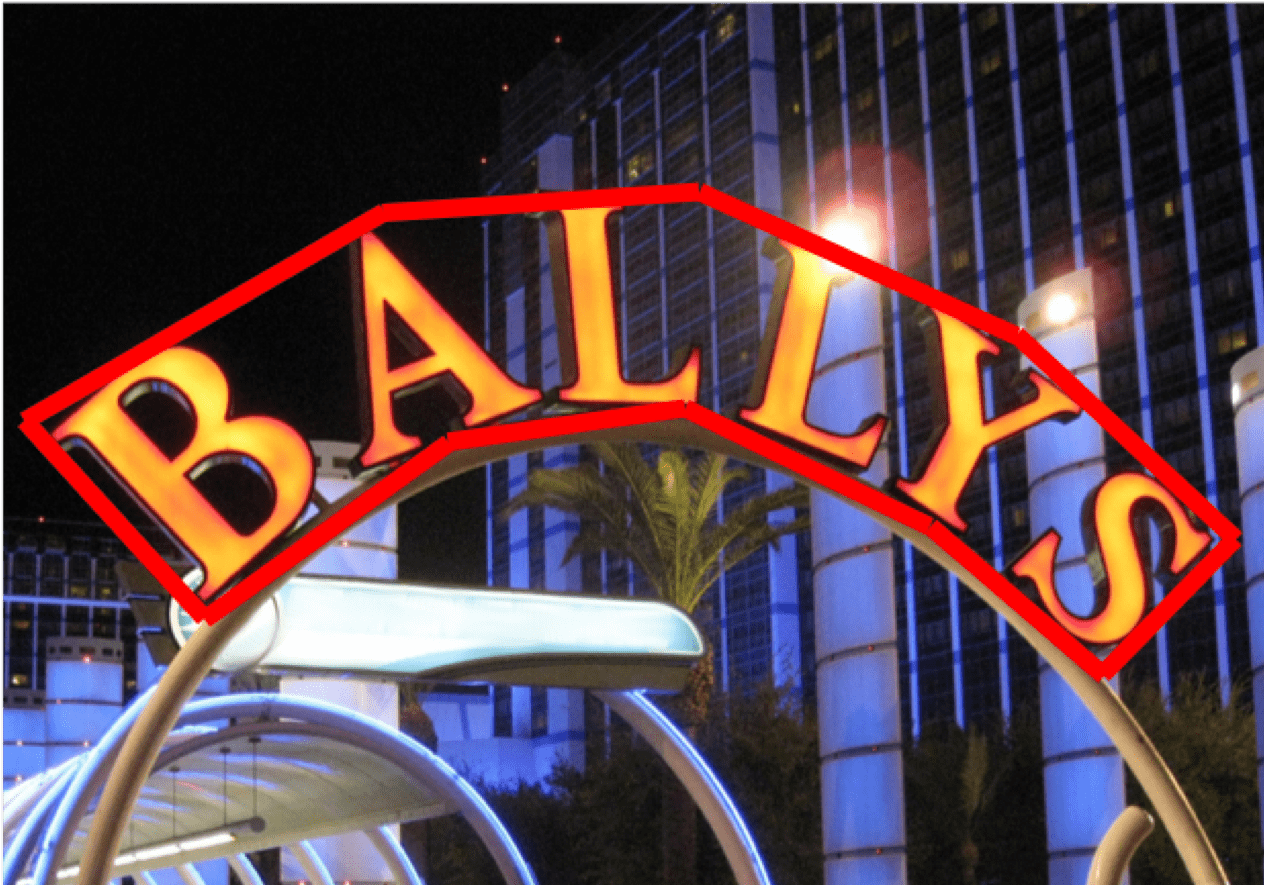}
  \caption{}
  \label{fig4:sfig1}
\end{subfigure}%
\begin{subfigure}{.16\textwidth}
  \centering
  \includegraphics[width=.95\linewidth]{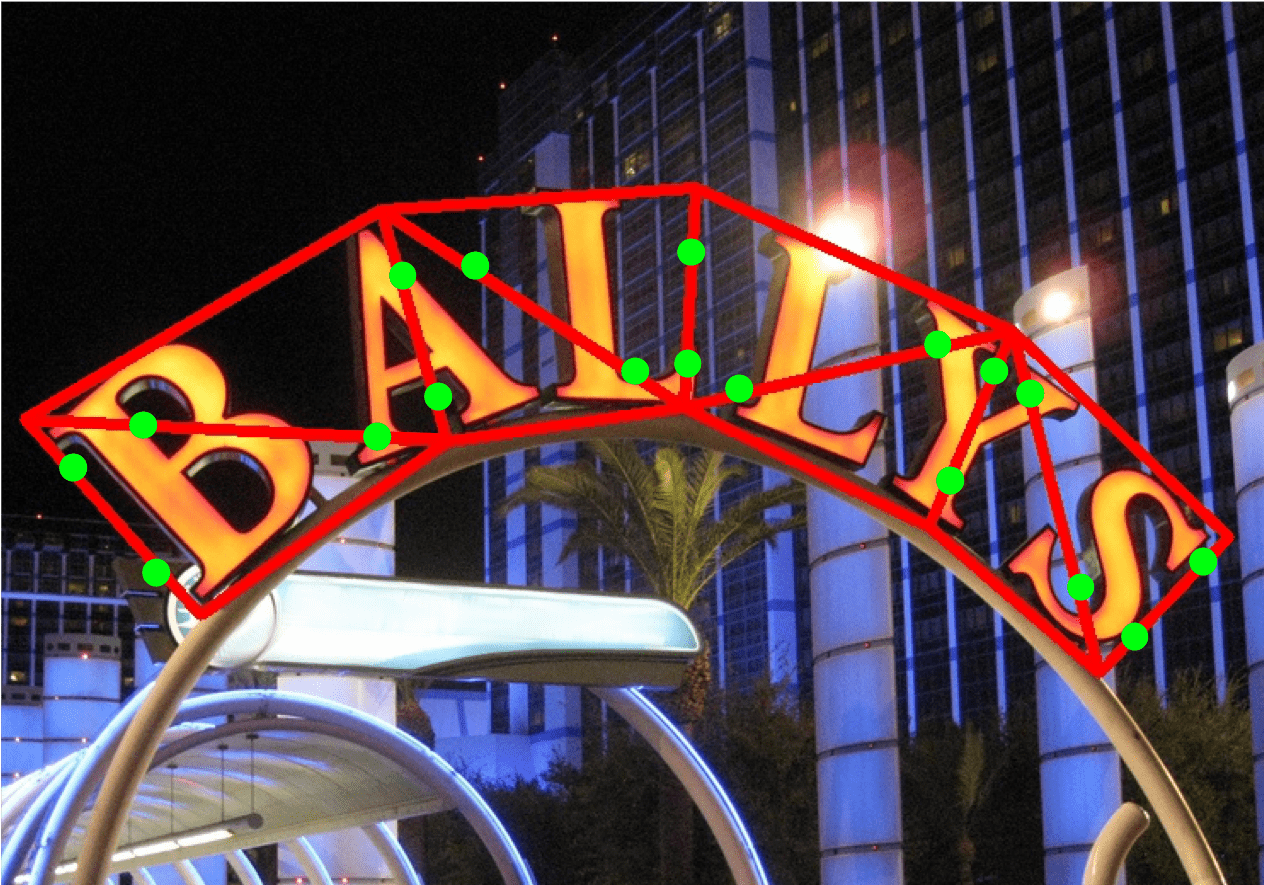}
  \caption{}
  \label{fig4:sfig2}
\end{subfigure}%
\begin{subfigure}{.16\textwidth}
  \centering
  \includegraphics[width=.95\linewidth]{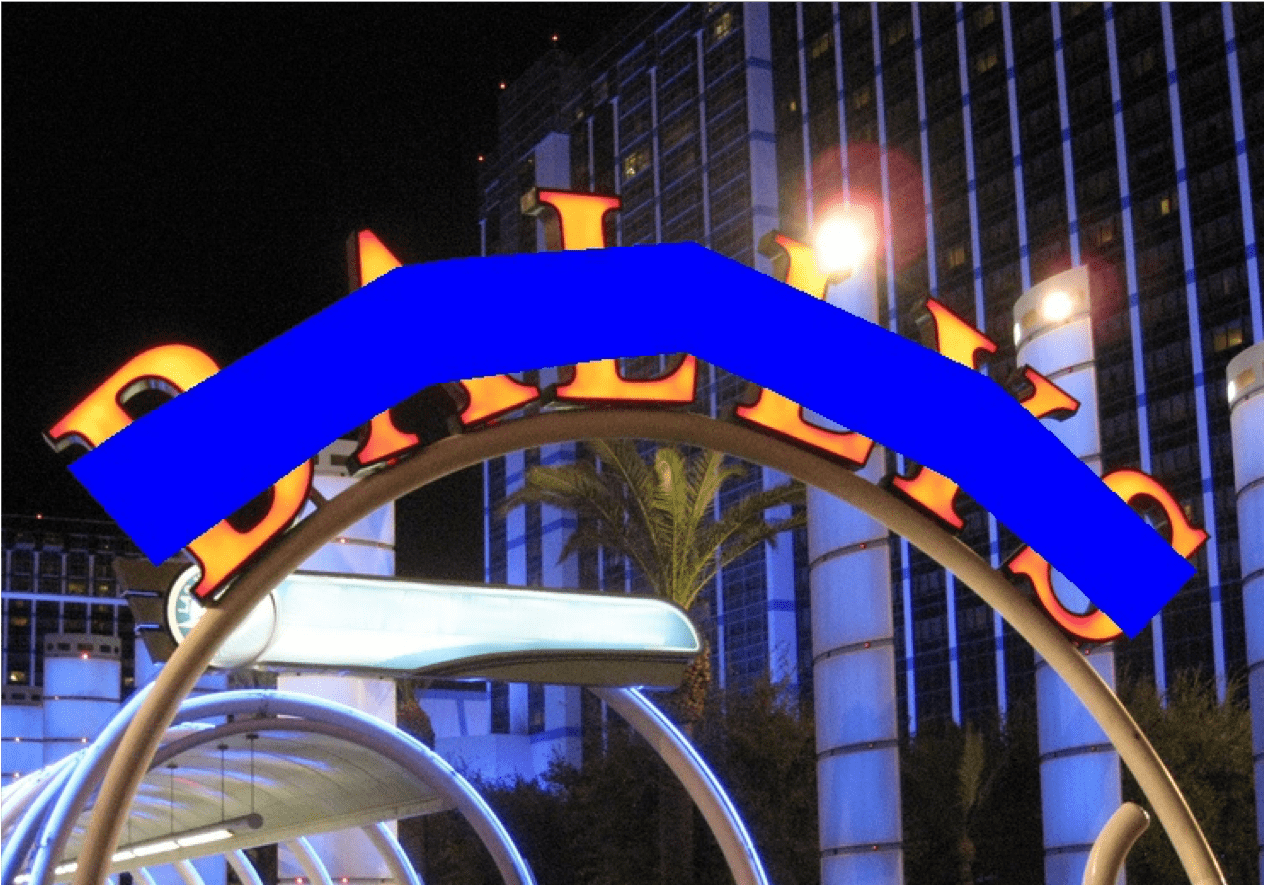}
  \caption{}
  \label{fig4:sfig3}
\end{subfigure}
\begin{subfigure}{.16\textwidth}
  \centering
  \includegraphics[width=.95\linewidth]{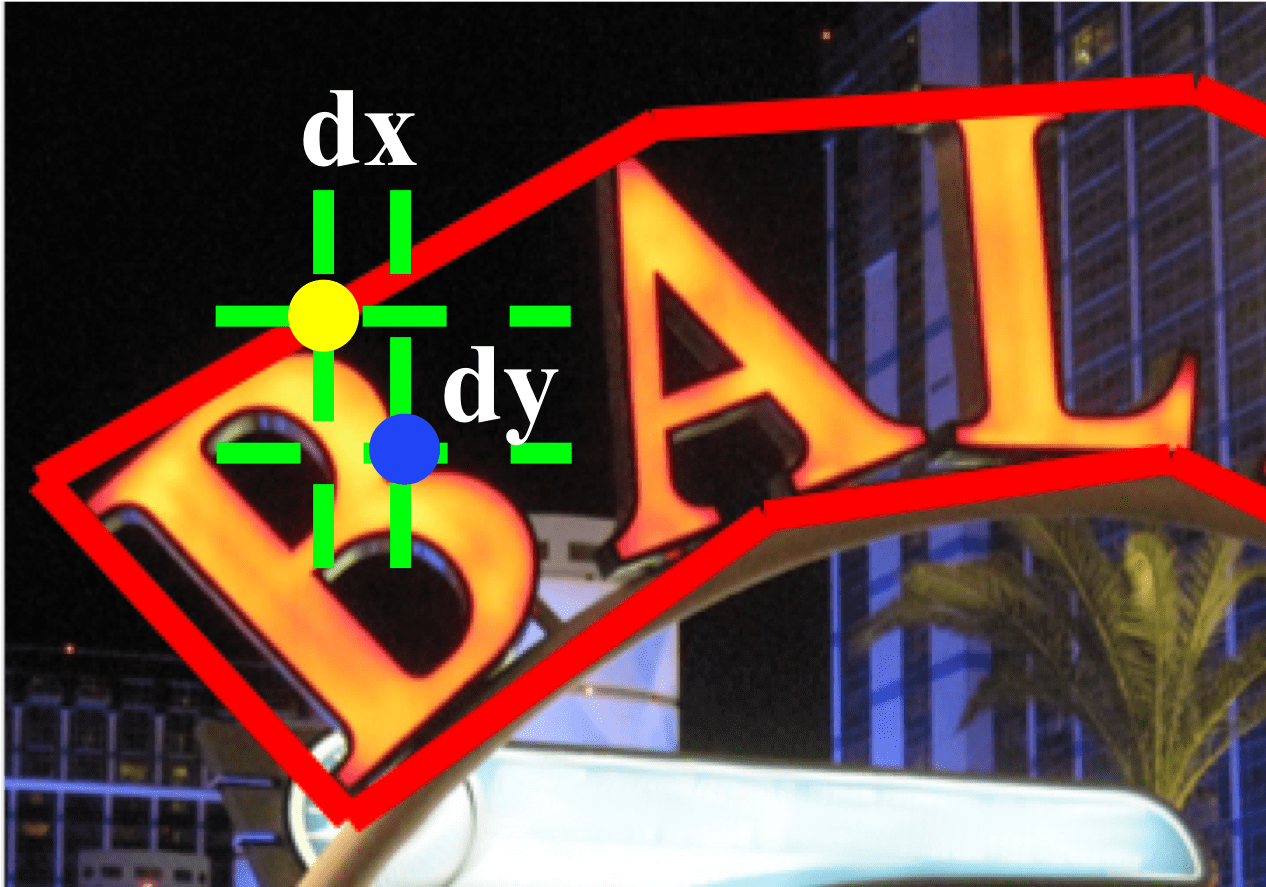}
  \caption{}
  \label{fig4:sfig4}
\end{subfigure}%
\begin{subfigure}{.16\textwidth}
  \centering
  \includegraphics[width=.95\linewidth]{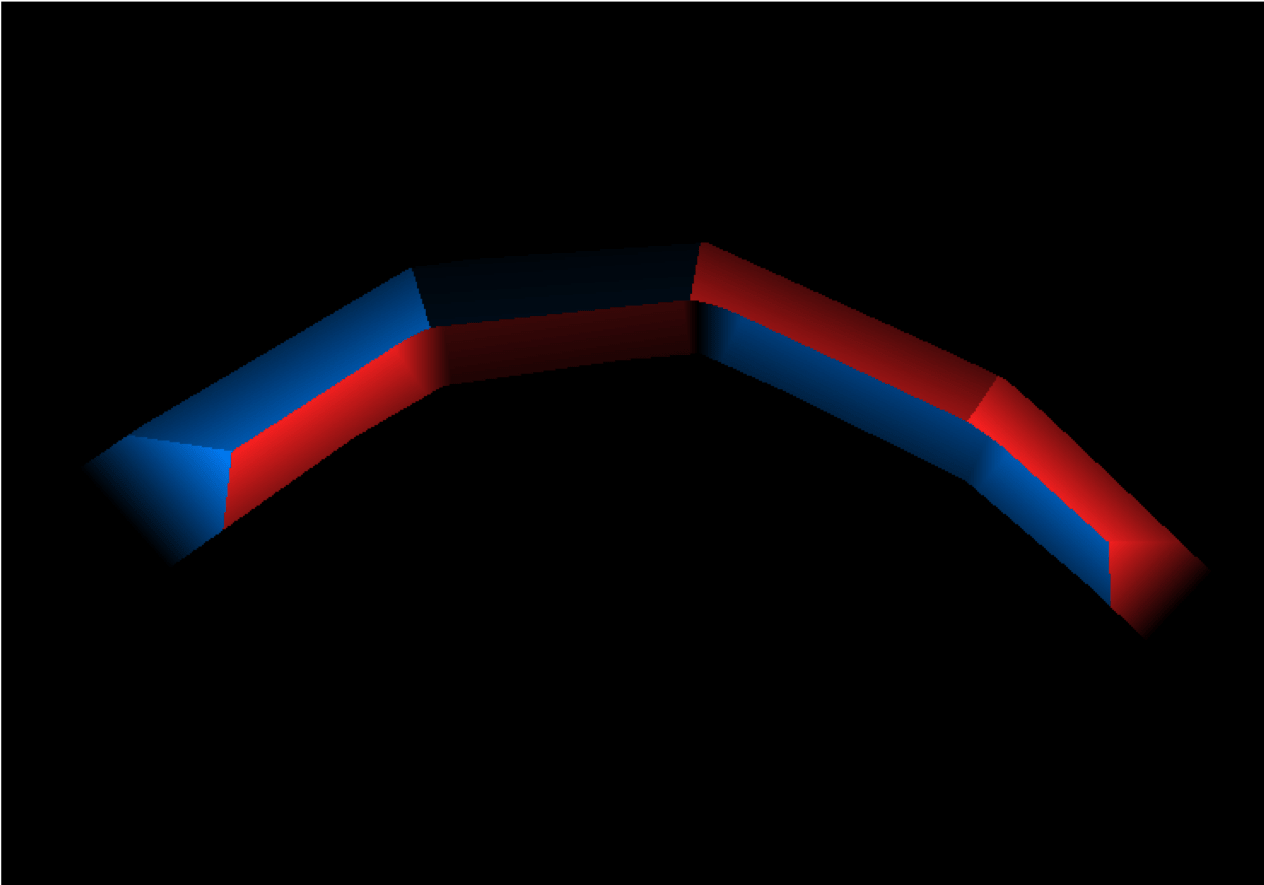}
  \caption{}
  \label{fig4:sfig5}
\end{subfigure}%
\begin{subfigure}{.16\textwidth}
  \centering
  \includegraphics[width=.95\linewidth]{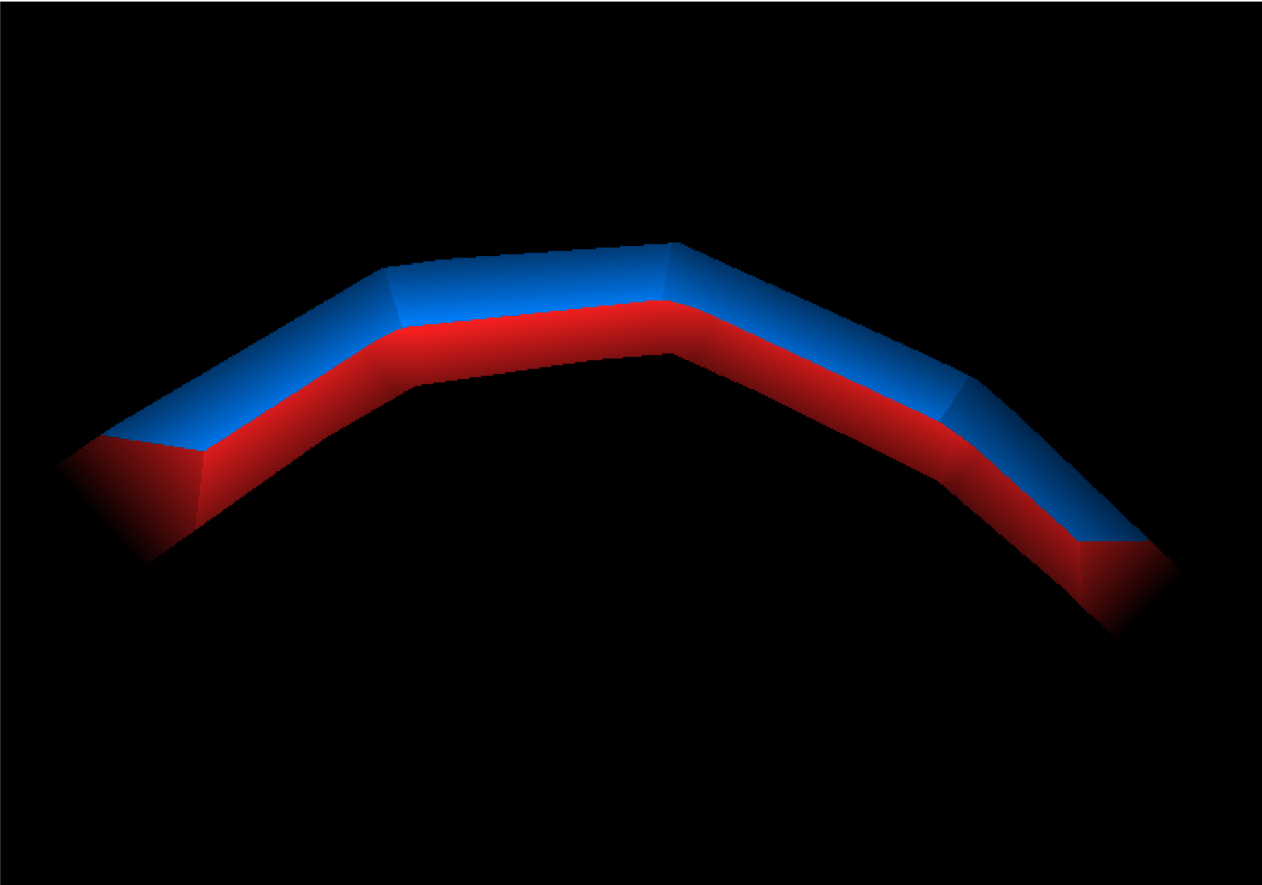}
  \caption{}
  \label{fig4:sfig6}
\end{subfigure}
\caption{Illustration of ground-truth generation: Given a text annotation polygon in (a), triangulation is performed over the polygon vertices to locate the vertices (green points in (b)) of the central text region in blue color in (c). For each central-text-region pixel $t_p$ (in blue color in (d)), the nearest point on the text annotation box $b_p$ in yellow color is determined as the nearest text boundary point as shown in (d), and the distance between $t_p$ and $b_p$ is used to generate ground-truth distance maps as shown in (e) and (f)
  }
\label{fig4}
\end{figure}

The proposed shape regression module first performs text pixel classification and regression. The classification predicts central text regions (as illustrated in the first graph under the \textbf{Shape Regression} in Fig. \ref{fig2}) by using the fused feature map from the \textbf{Multi-Scale Network}. The regression predicts two distance maps (as illustrated in the second and third graphs under the \textbf{Shape Regression}) according to the distance between each predicted text pixel and its nearest text boundary in horizontal and vertical directions (i.e. \textit{x} and \textit{y} coordinates), respectively. Note that the central text regions are derived from the original annotation boxes in training as illustrated in Figs. \ref{fig4:sfig1} and \ref{fig4:sfig3}. It is smaller than the annotation box which helps better separate neighboring words or text lines within the predicted text region map.

Each predicted text pixel will thus regress to one nearest point on text boundary which can be located by summing up the coordinates of the text pixel and the predicted distances in horizontal and vertical directions. Scene texts can thus be located by a polygon that encloses all detected text boundary points. We adopt the Alpha-Shape Algorithm \cite{akkiraju1995alpha} which produces a concave polygon enclosing a set of given points. In the Alpha-Shape Algorithm, triangle edges with radius larger than alpha thresholds are removed from the delaunay triangulation graph of the text boundary points. As the radius is sensitive to size of each triangle - i.e. size of text instance, the coordinates of boundary points of each text instance are first normalized to 0 to 1 to simplify the alpha threshold selection. The polygon for each text instance is therefore first generated from the normalized boundary points and then re-sized back to the original scale. 

To train the proposed shape regressor, a central text region and two distance maps are first extracted from the annotation polygon of each training text instance as illustrated in Fig. \ref{fig4}. Given a text annotation (may have four edges or more for curved text lines) as shown in Fig. \ref{fig4:sfig1}, triangulation is first performed over the annotation vertices, where each formed triangle has two vertices from the upper (or lower) side of the annotation and the third from the low (or upper) side as illustrated in Fig. \ref{fig4:sfig2}. For each newly formed triangle edge connecting the upper and lower sides of the text annotation, two points at 25\% distance from each end are determined which form the vertices of the central text region as illustrated in Figs. \ref{fig4:sfig2} and \ref{fig4:sfig3}. For each pixel $t_p$ in the central text region, the nearest point $b_p$ on the text annotation lines can be located (the yellow-color point in Fig. \ref{fig4:sfig4}), and the distances between $t_p$ and $b_p$ can then be determined to generate the distances maps in \textit{x} and \textit{y} directions as illustrated in Fig. \ref{fig4:sfig5} and \ref{fig4:sfig6}.

\begin{table*}[t]
\centering
\begin{tabular}{|c|c|c|c|c|c|c|}
\hline
                                         & \multicolumn{3}{c|}{\textbf{CTW1500}} & \multicolumn{3}{c|}{\textbf{Total-Text}} \\ \hline
\textbf{Methods}                         & \textbf{Precision} & \textbf{Recall} & \textbf{F-score} & \textbf{Precision}  & \textbf{Recall}  & \textbf{F-score} \\ \hline
\hline
SegLink \cite{shi2017detecting}          & 42.3      & 40.0   & 40.8    & 30.3       & 23.8    & 26.7    \\ \hline
CTD+TLOC* \cite{yuliang2017detecting}    & 77.4      & 69.8   & 73.4    & -          & -       & -       \\ \hline
Mask TextSpotter* \cite{Lyu_2018_ECCV}   & -         & -      & -       & 69.0       & 55.0    & 61.3    \\ \hline
TextSnake* \cite{long2018textsnake}      & 67.9 & \textbf{85.3} & 75.6  & 82.7       & 74.5    & 78.4    \\ \hline
\hline
\textbf{EAST(Baseline)} \cite{zhou2017east} & 78.7      & 49.1   & 60.4    & 50.0       & 36.2    & 42.0    \\ \hline
\textbf{Ours*}                           & \textbf{85.0} & 78.3 & \textbf{81.5} & \textbf{83.8} & \textbf{74.8} & \textbf{79.0} \\ \hline            
\end{tabular}
\caption{Experimental results over the curved-text-line datasets \textbf{CTW1500} and \textbf{Total-Text}, where methods with `*' specifically address curved text lines. Results of SegLink and EAST are taken from \protect\cite{long2018textsnake}.}
\label{tab:curve}
\end{table*}

\subsection{Multi-Scale Multi-Stage Detection Network}
The proposed network adopts a multi-channel structure to accommodate images of different scales. Given a training image, it is first re-sampled by a factor of 2 and produces multiple re-scaled images. The training image and the re-scaled are then fed to multiple network channels for feature extraction. Fig. \ref{fig3} shows one network structure that uses two channels for the original training image and the half-scaled (this network is adopted in our implemented scene text detection system). Within each network channel, image features are extracted at multiple network stages to capture details at different levels. At the end, object features extracted from multiple network channels and multiple network stages (within each channel) are fused and fed to the \textbf{Shape Regression} module which will predict the central text regions and distance maps.

The simultaneous feature learning from images of different scales at different network stages is a challenging task as features from the same network stage of different network channels have different scales. Fig. \ref{fig3} illustrates how our proposed network architecture addresses this challenge. As Fig. \ref{fig3} shows, features from \textit{Conv5} of \textit{Channel 2} are firstly up-sampled by a factor of 2 so that they will have the same scale as features from \textit{Conv5} of \textit{Channel 1} (and so \textit{Con4} of \textit{Channel 2}). A \textit{Concat UpConv} module is exploited for feature fusion, which first concatenates features from three network stages and then up-samples the concatenated features by a factor of 2 (after a 1x1 convolution and a 3x3 convolution) \cite{zhou2017east} and finally passes the up-sampled feature to the earlier network stage for further feature fusion. Note we only use features from stages from \textit{Conv2} to \textit{Conv5} as those from \textit{Conv1} are too low-level.

The proposed network improves the detection of scene texts of very different sizes from two aspects. On the one hand, the multi-stage design improves the prediction of details at different levels by fusing local and global features effectively. On the other hand, the multi-scale design extracts features from text images of different scales which addresses the large text size variation directly.
\begin{table*}[t]
\centering
\begin{tabular}{|c|c|c|c|c|c|c|c|}
\hline
\textbf{}                                   & \multicolumn{4}{c|}{\textbf{ICDAR2015}}                      & \multicolumn{3}{c|}{\textbf{MSRA-TD500}}      \\ \hline
\textbf{Methods}                            & \textbf{Prediction}    & \textbf{Recall}    & \textbf{F-score}    & \textbf{FPS} & \textbf{Prediction}    & \textbf{Recall}    & \textbf{F-score}    \\ \hline
\hline
SegLink \cite{shi2017detecting}             & 73.1          & 76.8          & 75.0          & -            & 86.6          & 70.0          & 77.0          \\ \hline
PixelLink \cite{deng2018pixellink}          & 85.5          & \textbf{82.0} & \textbf{83.7} & 3.0          & 83.0          & 73.2          & 77.8          \\ \hline
TextSnake* \cite{long2018textsnake}         & 84.9          & 80.4          & 82.6          & 1.1          & 83.2          & 73.9          & 78.3          \\ \hline
Mask TextSpotter* \cite{Lyu_2018_ECCV}      & 85.8          & 81.2          & 83.4          & 4.8          & -             & -             & -             \\ \hline
RRD \cite{liao2018rotation}                 & 85.6          & 79.0          & 82.2          & \textbf{6.5}          & 87.0          & 73.0          & 79.0          \\ \hline
Lyu \textit{et al.} \cite{lyu2018multi}     & \textbf{94.1} & 70.7          & 80.7          & 3.6          & \textbf{87.6}          & 76.2          & 81.5          \\ \hline
\hline
\textbf{EAST(Baseline)} \cite{zhou2017east} & 80.5          & 72.8          & 76.4          & \textbf{6.5}          & 87.3          & 67.4          & 76.1          \\ \hline
\textbf{Ours*}                              & 86.6          & 78.4          & 82.3          & 4.3          & 87.4          & \textbf{76.7}          & \textbf{81.7} \\ \hline
\end{tabular}
\caption{Experimental results over the straight-text-line datasets \textbf{ICDAR2015} and \textbf{MSRA-TD500}, where methods with `*' specifically address curved text lines. Only single-scale testing results are listed for ICDAR2015 dataset for fair comparison.
}
\label{tab:quad}
\end{table*}

\subsection{Network Training}
The proposed multi-scale shape regression network takes the original training image and the corresponding central text regions and distances maps as inputs. The training aims to minimize the following multi-task loss function:

\begin{equation}
\mathcal{L}=\mathcal{L}_{cls} + \lambda*\mathcal{L}_{reg}
\end{equation}
\noindent where $\mathcal{L}_{cls}$ and $\mathcal{L}_{reg}$ refer to the loss of classification (for prediction of central text regions) and regression (for prediction of distances to the nearest text boundary), respectively. Parameters $\lambda$ is the weight to balance the two losses which is empirically set at 1.0 in our implemented system.

The prediction of central text regions is actually a pixel-wise binary classification problem. We adopt the Dice Coefficient loss that is defined by:
\begin{equation}
\mathcal{L}_{cls}=\frac{2*|G\cap P|}{|G|+|P|}
\end{equation}
\noindent where \textit{G} and \textit{P} refer to the ground-truth central text region and the predicted central text region, respectively.

The prediction of distances from central text pixels to the nearest text boundary is a regression problem. We define the regression loss based on the Smooth L1 loss \cite{girshick2015fast}: 
\begin{equation}
\begin{split}
\mathcal{L}_{reg}=0.5*\sum_{k}Smooth_{L1}(x_{k}-x_{k}^*) \\ 
+0.5*\sum_{k}Smooth_{L1}(y_{k}-y_{k}^*)
\end{split}
\end{equation}
\noindent where \textit{x} and \textit{y} denote ground-truth distances in the horizontal and vertical directions, respectively, and \textit{$x^*$} and \textit{$y^*$} denote the correspondingly predicted distances. The $Smooth_{L1}$ denotes the standard Smooth L1 loss.

\section{Experiments}
\subsection{Datasets}
\medskip \noindent 
\textbf{SynthText} \cite{Gupta16} contains more than 800,000 synthetic scene text images most of which are at word level with multi-oriented rectangular annotations.

\medskip \noindent 
\textbf{CTW1500} \cite{yuliang2017detecting} consists of 1,000 training images and 500 test images that contain 10,751 multi-oriented text instances of which 3,530 are arbitrarily curved. Each text instance is annotated at text-line level by using 14 vertices, where texts are largely in English and Chinese. 

\medskip \noindent 
\textbf{Total-Text} \cite{ch2017total} has 1,255 training images and 300 test images where texts are all in English. It contains a large number of multi-oriented curved text instances each of which is annotated at word level with a polygon. 

\medskip \noindent 
\textbf{MSRA-TD500} \cite{yao2012detecting} consists of 300 training images and 200 test images. All captured text instances are printed in English and Chinese which are annotated at text-line level by using best-alighed rectangles. 

\medskip \noindent 
\textbf{ICDAR2015} \cite{karatzas2015icdar} has 1000 training images and 500 test images which are collected by Google Glass and suffers from low resolution and motion blur. All text instances are annotated at word level using quadrilateral boxes.

\begin{figure*}[t!]
  \centering
  \includegraphics[width=\linewidth]{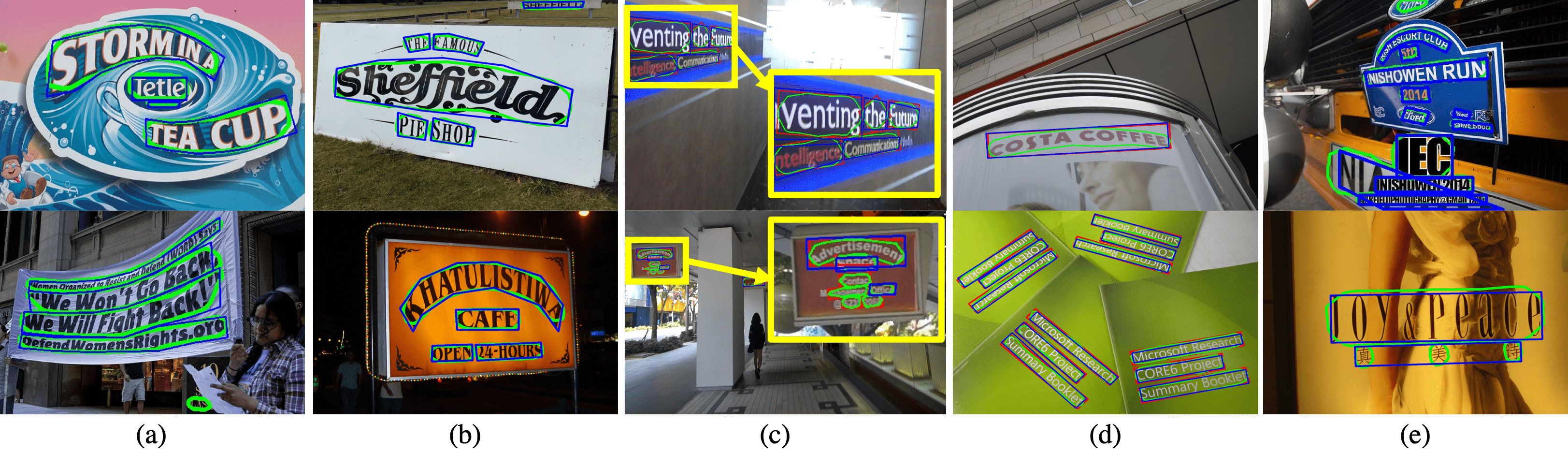}
  \caption{Illustration of the proposed scene text detection method: Sample images in (a)-(d) are selected from CTW1500, Total-Text, ICDAR2015 and MSRA-TD500 respectively, where green boxes show polygon localization boxes by our proposed method and blue boxes show ground-truth boxes. For images with straight text lines in (c) and (d), red quadrilateral boxes are derived from the polygon localization boxes for the evaluation purpose. A few typical unsuccessful cases are given in (e).  }
  \label{fig5}
\end{figure*}

\subsection{Implementation Details}
The proposed technique is implemented using Tensorflow on a regular GPU workstation with 2 Nvidia Geforce GTX 1080 Ti. The network is optimized by Adam optimizer \cite{kingma2014adam} with a starting learning rate of $10^{-4}$. The network is pre-trained on the SynthText, which is then fine-tuned by using the training images of each evaluated dataset with a batch size of 10. ResNet-50 \cite{he2016deep} is used as the network backbone.

\subsection{Experimental Results}
\begin{figure*}[t!]
  \centering
  \includegraphics[width=\linewidth]{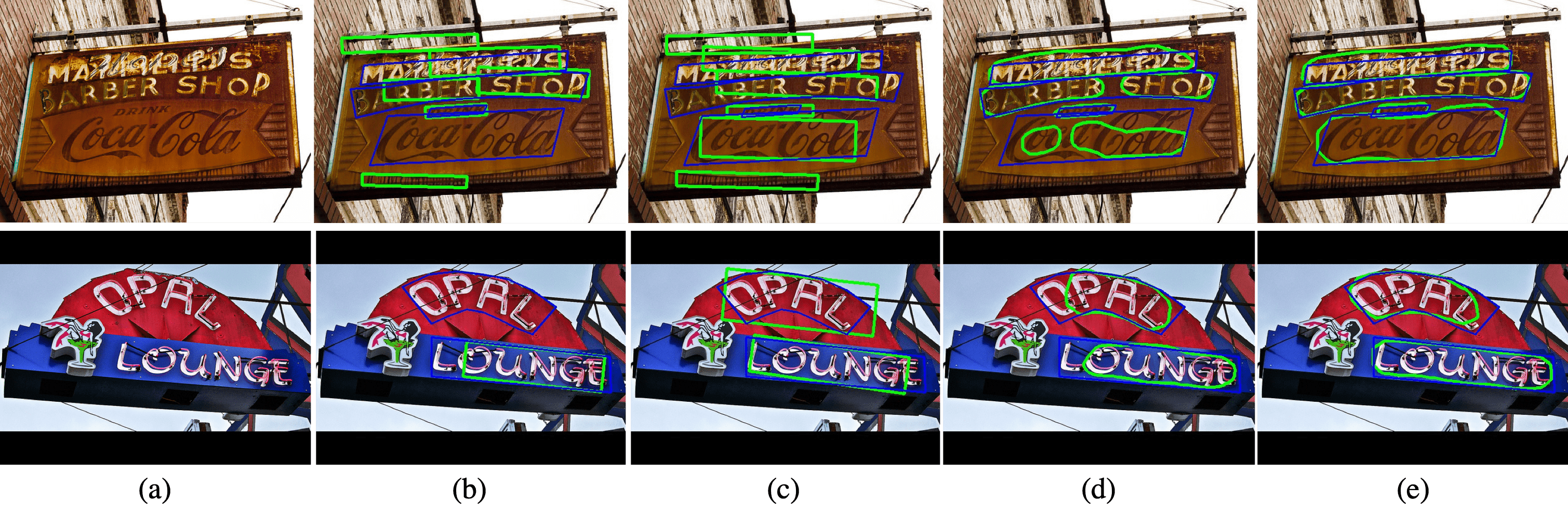}
  \caption{Ablation study of the proposed technique: For sample images from CTW1500 in (a), the green-color boxes in (b-e) show scene text detection by using the \textbf{`Baseline'}, \textbf{`Baseline+Multi-Scale'}, \textbf{`Baseline+Shape Regression'} and \textbf{`Multi-Scale Shape Regression'}, respectively. The ground truth are shown in blue-color boxes.
  }
  \label{fig6}
\end{figure*}

The proposed technique has been evaluated quantitatively and qualitatively over four public datasets as shown in Tables \ref{tab:curve} and \ref{tab:quad} and Fig. \ref{fig5}. It has also been analyzed through ablation studies as shown in Table \ref{tab:ablation} and Fig. \ref{fig6}.

\subsubsection{Texts in Different Orientations and Shapes}
The proposed technique has been evaluated over the datasets CTW1500 and Total-Text where many scene texts were captured in different orientations, shapes and lengths. The purpose is to study how the proposed technique performs under the presence of many different text appearances. 

As Table \ref{tab:curve} shows, the proposed method achieves f-scores of 81.5\% and 79.0\% on datasets CTW1500 and Total-Text in single-scale testing mode which are significantly higher than state-of-the-art methods that didn't specifically address curved text lines. In particular, it outperforms the best f-score by 5.9 for CTW1500 with annotations at text-line level, demonstrating its superiority in dealing with text lines of different lengths. In fact, the proposed method is capable of dealing with text lines of very different lengths as illustrated in Fig. \ref{fig5} because it regresses to the nearest text boundary instead of the four quadrilateral vertices. For Total-Text with scene texts annotated at word level, our proposed method also achieved state-of-the-art performance, demonstrating its superiority in dealing with curved texts with limited length variations. In addition, the proposed method is on par with or significantly outperform the very recent methods that specifically addressed curved text lines. 

\subsubsection{Texts in Different Orientations and Lengths}
The proposed method is also evaluated over ICDAR2015 and MSRA-TD500 where most scene texts are straight but in different orientations with annotations at word and text-line level. We derive an oriented rectangular box from each determined concave polygon as illustrated in the second row of Fig. \ref{fig5} (highlighted in red-color boxes) for overlap computation in evaluations. Table \ref{tab:quad} shows experimental results and comparisons with state-of-the-art techniques. 

For MSRA-TD500, the proposed method achieves state-of-the-art f-score, demonstrating it superior capability in dealing with straight text lines of different orientations and lengths, the subject that has been studied for many years in the scene text detection community. In fact, it outperforms most state-of-the-art methods that predict rectangular boxes which are more suitable for straight text lines. Further, it outperforms TextSnake, a very recent method that specifically addressed curved text lines, by up to 3.4 in f-score. For ICDAR2015, our approach achieves competitive performance in both accuracy and speed at single-scale testing mode.

\subsection{Ablation Study}
The proposed multi-scale shape regression network consists of two innovative components, namely, a multi-scale network and a shape regression module. We perform an ablation study over CTW1500 to identify the contribution of these two components. Four models are trained as shown in Table \ref{tab:ablation}. The first is \textbf{`Baseline'} which refers to the original EAST model \cite{zhou2017east} that regresses text pixels to four quadrilateral vertices. The second is \textbf{`Baseline+Multi-Scale'} which uses EAST but includes the proposed multi-scale network structure. The third is \textbf{`Baseline+Shape Regression'} that uses EAST but regresses to the nearest text boundary. The last is \textbf{`Multi-Scale Shape Regression'} that fully implements both \textbf{Multi-Scale} and \textbf{Shape Regression}. \footnotebl{Acknowledgement: This work is funded by Ministry of Education, Singapore, under the project ``A semi-supervised learning approach for accurate and robust detection of texts in scenes" (RG128/17 (S)).}

As Table \ref{tab:ablation} shows, the inclusion of multi-scale network alone improves the recall significantly with certain sacrifice of precision, and the inclusion of the shape regression alone improves both recall and precision clearly, leading to a 17 improvement in f-score. In addition, the inclusion of both multi-scale network and shape regression module improves the f-score by over 21 beyond the baseline. Fig. \ref{fig6} illustrate the ablation study where many missing and broken detection are correctly detected by the \textbf{Multi-Scale+Shape Regression}.
\begin{table}[t]
\centering
 \begin{tabular}{|l | c | c | c|} 
 \hline
 \textbf{Methods} & \textbf{P} & \textbf{R} & \textbf{F-score} \\ [0.5ex] 
 \hline\hline
 Baseline & 78.7 & 49.1 & 60.4 \\
 Baseline+Multi-Scale & 72.8 & 60.8 & 66.3 \\
 Baseline+Shape Regression & 82.8 & 72.1 & 77.1 \\
 Multi-Scale Shape Regression & \textbf{85.0} & \textbf{78.3} & \textbf{81.5} \\
 \hline
 \end{tabular}
 \caption{Ablation study of the proposed technique over the dataset CTW1500 (P: precision; R: recall)}
 \label{tab:ablation}
\end{table}

\subsection{Discussion}

The proposed method still faces certain constraints under several specific scenarios. First, it could fail while dealing with text lines that spatially overlap with each other as shown in the first row of Fig. \ref{fig5} (e), largely due to the ambiguity in differentiating the central text region while text lines overlap with each other. Second, it could produce broken detection when characters in a word or text line are widely separated as shown in the last example image in the second row of Fig. \ref{fig5} (e). Without text semantics, it's a common challenge to decide whether characters/letters belong to the same text line when they are widely separated. 

\section{Conclusion}
This paper presents a novel multi-scale shape regression network that is capable of locating scene texts of different orientations, shapes and lengths accurately. The proposed method predicts dense text boundary points instead of sparse quadrilateral vertices that are prone to produce large regression errors while dealing with long text lines. In addition, this also enables accurate localization of scene texts of arbitrary orientations and curvatures whereas state-of-the-art techniques using quadrilaterals often include undesired background to the ensuing scene text recognition task. The multi-scale network extracts and fuses features at different scales which demonstrates superb tolerance to the text scale variation. Extensive experiments over several public datasets show the superior performance of the proposed technique.

\bibliographystyle{named}
\bibliography{ijcai19}

\end{document}